%% file: main.tex
\documentclass[sigconf]{acmart}

\usepackage{lipsum} 
\usepackage{comment} 
\usepackage{xcolor}
\usepackage{xspace}
\usepackage{microtype}
\usepackage{graphicx}
\usepackage{subcaption}
\usepackage{booktabs}
\usepackage{tabularx}
\usepackage{soul}
\usepackage{hyperref}
\usepackage[framemethod=tikz]{mdframed}
\usepackage{marvosym}
\usepackage{listings} 

\makeatletter
\def\@ACM@checkaffil{
    \if@ACM@instpresent\else
    \ClassWarningNoLine{\@classname}{No institution present for an affiliation}%
    \fi
    \if@ACM@citypresent\else
    \ClassWarningNoLine{\@classname}{No city present for an affiliation}%
    \fi
    \if@ACM@countrypresent\else
        \ClassWarningNoLine{\@classname}{No country present for an affiliation}%
    \fi
}
\makeatother

\setcopyright{acmcopyright}
\copyrightyear{2023}
\acmYear{2023}
\acmDOI{XXXXXXX.XXXXXXX}
\acmConference[MLBench'23]{Workshop on Benchmarking Machine Learning Workloads on Emerging Hardware}{June 8, 2023}{Miami, FL}
\acmPrice{15.00}
\acmISBN{978-1-4503-XXXX-X/18/06}

\newcommand{\Chakraet}{\texttt{Chakra ET}\xspace}

\newcommand*\circled[1]{\tikz[baseline=(char.base)]{\node[shape=circle,fill,inner sep=0.7pt] (char) {\textcolor{white}{#1}};}}

\input{authors}

\begin{document}

\title{Chakra: Advancing Performance Benchmarking and Co-design using Standardized Execution Traces}

\input{abstract}

\maketitle
\pagestyle{plain}

\input{introduction}
\input{overview}
\input{schema}
\input{generative_model}
\input{open_source_tools}
\input{use_cases}
\input{related_work}
\input{conclusion}
\input{acknowledgements}

\bibliographystyle{ACM-Reference-Format}
\bibliography{references}

\end{document}

%% file: authors.tex
\author{Srinivas Sridharan\textsuperscript{1\Cross}, Taekyung Heo\textsuperscript{1, 2\Cross}, Louis Feng\textsuperscript{1}, Zhaodong Wang\textsuperscript{1}, Matt Bergeron\textsuperscript{1}, Wenyin Fu\textsuperscript{1}, Shengbao Zheng\textsuperscript{1}, Brian Coutinho\textsuperscript{1}, Saeed Rashidi\textsuperscript{2, 3\ddag}, Changhai Man\textsuperscript{2}, Tushar Krishna\textsuperscript{2}}

\affiliation{
  \textsuperscript{1}Meta
}
\affiliation{
  \textsuperscript{2}Georgia Institute of Technology
}
\affiliation{
  \textsuperscript{3}Hewlett Packard Enterprise
}

%% file: abstract.tex
\begin{abstract}
\label{sec:abstract}
Benchmarking and co-design are essential for driving optimizations and innovation around ML models, ML software, and next-generation hardware. Full workload benchmarks, e.g. MLPerf, play an essential role in enabling fair comparison across different software and hardware stacks especially once systems are fully designed and deployed. However, the pace of AI innovation demands a more agile methodology to benchmark creation and usage by simulators and emulators for future system co-design. We propose Chakra\footnote{https://github.com/chakra-et/chakra}, an open graph schema for standardizing workload specification capturing key operations and dependencies, also known as Execution Trace (ET). In addition, we propose a complementary set of tools/capabilities to enable collection, generation, and adoption of Chakra ETs by a wide range of simulators, emulators, and benchmarks. For instance, we use generative AI models to learn latent statistical properties across thousands of Chakra ETs and use these models to synthesize Chakra ETs. These synthetic ETs can obfuscate key proprietary information and also target future what-if scenarios. As an example, we demonstrate an end-to-end proof-of-concept that converts PyTorch ETs to Chakra ETs and uses this to drive an open-source training system simulator (ASTRA-sim). Our end-goal is to build a vibrant industry-wide ecosystem of agile benchmarks and tools to drive future AI system co-design.
\end{abstract}

%% file: introduction.tex
\section{Introduction}
\label{sec:introduction}
Modern deep learning systems heavily rely on distributed training over high-performance \emph {Neural Processing Unit} (NPUs) (i.e., GPU or TPU)-based hardware platforms (e.g., Meta ZionEX, Google CloudTPU, NVIDIA DGX). Distributing the training task involves splitting the datasets (data parallel) or splitting the model weights (model parallel) or splitting the model layers (pipeline parallel) or hybrid combinations (3D/4D parallel). The optimal parallelism strategy is a topic of active research~\cite{jia2019beyond, rajbhandari2020zero}.

\def\thefootnote{\Cross}\footnotetext{These authors contributed equally to this work}
\def\thefootnote{\ddag}\footnotetext{This work was done while the author was affiliated with Georgia Tech}

There is an agreement in the community that benchmarking and HW-SW co-design are crucial for developing next-generation AI/ML platforms. Full workload benchmarks, e.g. MLPerf, play an essential role in enabling fair comparison across different software and hardware stacks especially once systems are fully designed and deployed. However, the pace of AI innovation demands a more agile methodology for benchmark creation and usage by simulators and emulators for future system co-design.

Furthermore, state-of-the-art AI/ML models and custom AI/ML accelerators (NPUs) are often designed by different organizations (e.g., DLRM designed by Meta running on NVIDIA GPUs) – and these organizations (especially industry) cannot disclose full details due to proprietary intellectual properties (IPs) and technologies. As a result, hyperscalar cloud providers today end up sharing spreadsheets with model parameters under NDA with some vendors, but it is hard to derive/reproduce exact workload details; while the majority of the hardware vendors (and academic researchers) have to derive parameters from MLPerf or other benchmarks, often over-optimize a few use cases and ignoring Amdahl's law.

To overcome these challenges, this paper proposes Chakra, an infrastructure for enabling benchmarking of ML models on \textit{future} systems and performing co-design-space exploration. The key enabler of Chakra is the idea of \textit{execution traces (ET)}. Analogous to instruction or memory traces~\cite{google_workload_traces} for optimizing compute and memory architectures, ETs encode critical information related to compute and communication operator dimensions and dependencies while not revealing model or dataset details. SW companies can share ET for their internal workloads of interest to HW vendors who can in-turn estimate and optimize performance on next-generation NPU platforms using their proprietary HW model/simulator(s). The feedback can help SW companies identify optimized compute and network configurations for their training tasks. The availability of ETs can also help academic researchers and HW startups participate in co-design for real production ML workloads.

The information required by ET is in fact readily available during execution of a ML workload using a framework like PyTorch/TensorFlow, but not explicitly exposed today. This paper presents the key components of the Chakra infrastructure - (i) an ET schema, called \Chakraet that captures relevant execution and dependency information\footnote{We plan to standardize this via MLCommons, OCP, or other existing standards organization in consultation with other hyperscalars and hardware vendors.}, (ii) open-source toolchains such as visualizers, converter, and simulators, and (iii) generative AI model for synthesizing execution traces.

%% file: overview.tex
\section{Chakra Overview}
\label{sec:chakra}
\begin{figure}[t]
    \centering
    \includegraphics[width=.98\linewidth]{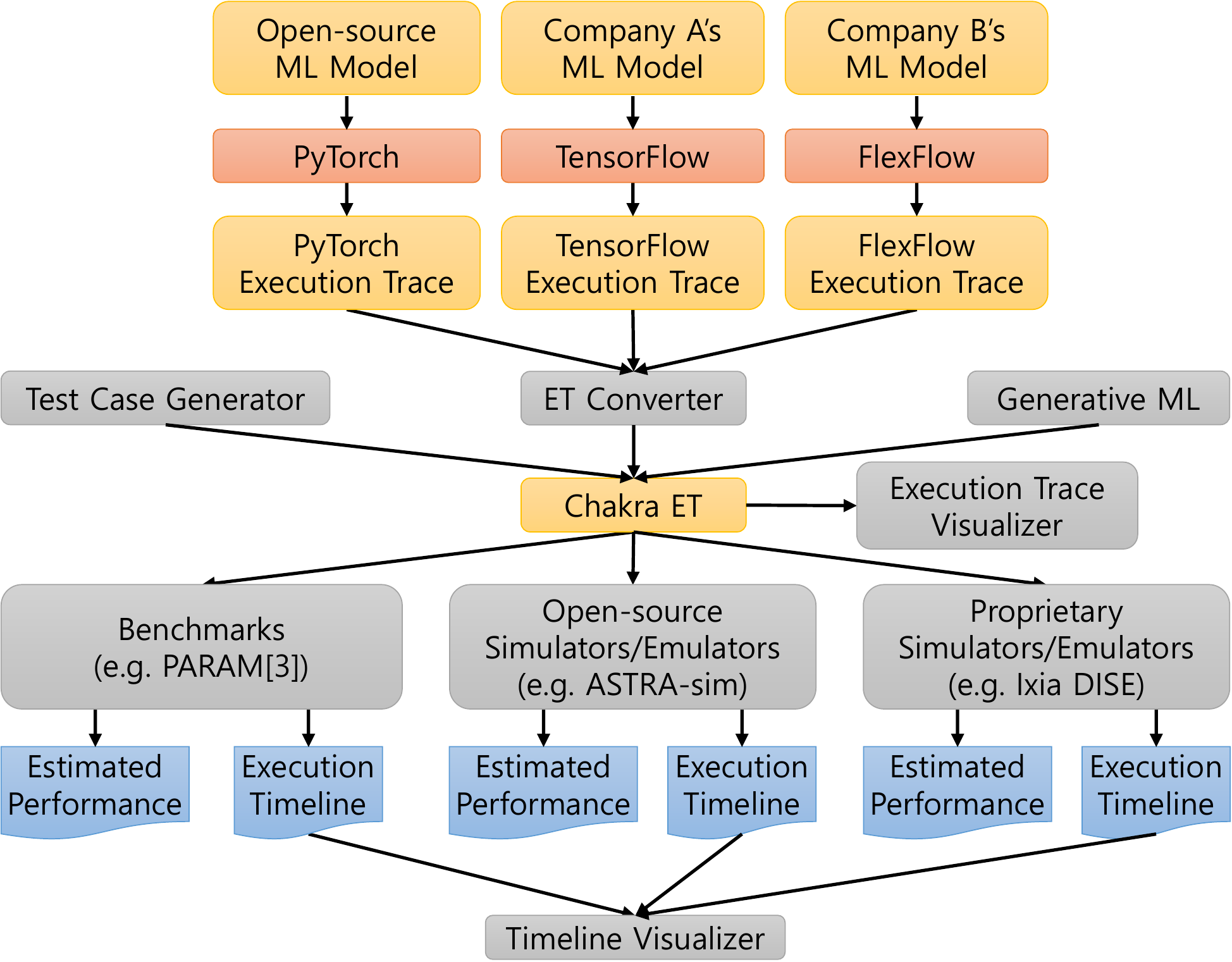}
    \caption{Chakra infrastructure overview.}
    \label{fig:Chakra_infrastructure_overview}
\end{figure}

In this section, we overview Chakra, a performance modeling framework for distributed ML workloads. Comprising various concepts and tools, Chakra enables users to estimate the execution time and resource usage of a distributed ML task for a given system configuration. \autoref{fig:Chakra_infrastructure_overview} illustrates the key components of the Chakra infrastructure, with execution traces (ETs) serving as the central element. These ETs facilitate the exchange of ML execution traces without revealing model specifics. Generated by ML frameworks, ETs are converted to a common format, \Chakraet, which are subsequently fed into simulators to estimate the performance of ML workloads. In this section, we provide an overview of the Chakra project and its main components.

\subsection{Motivation}
\label{sec:motivation}
The importance of HW-SW codesign in designing distributed ML systems is well-known. Despite its importance, the necessary infrastructure for effective implementation remains insufficient. Three primary challenges hinder progress in this area. Firstly, the absence of a unified schema for exchanging execution traces of ML models poses a significant obstacle. Although numerous ML frameworks and languages exist, their primary focus has been on representing and optimizing ML models, with performance modeling receiving less attention. Consequently, no standard schema for encoding execution information has been developed. Secondly, the lack of toolchains for performance modeling is another challenge. Comprehensive toolchains are crucial for identifying bottlenecks and debugging issues efficiently. Lastly, current methodologies lack the ability to synthesize execution traces. Synthesizing traces is vital for several reasons, including concerns over sharing intellectual property when exchanging execution traces between different organizations. By generating synthetic traces that retain the characteristics of real-world data while obscuring sensitive details, companies can share information without compromising their proprietary knowledge. Furthermore, synthesized execution traces play a critical role in performance projection, as future systems will likely have varying numbers of NPUs and connectivity configurations.

\subsection{Schema}
In Chakra, execution traces are in the Chakra schema, which is defined as a common format for exchanging execution traces for performance modeling purposes. The need for a standardized schema arises as ML frameworks often provide methods to collect execution traces of ML models, but these traces are presented in varying formats, which obstructs the exchange of information between different organizations. To address this issue, the Chakra schema is designed based on the requirements of various organizations, including universities and companies. Moreover, the schema facilitates the collection of traces at different stages, such as pre-execution and post-execution, allowing for the gathering of traces that are either specific to a particular system design or not.

\subsection{Chakra ET Collection and Synthesis}
\label{sec:et-sources}
Execution traces can be collected from real-world ML frameworks or synthetic trace generators, each with its own pros and cons. ML frameworks offer the most straightforward method for collecting execution traces, as they run ML models directly. We extended PyTorch to support trace collection with the Execution Graph Observer~\cite{pytorch_eg_observer}, which does not require intrusive modifications to ML models. Simply enabling the Execution Graph Observer before running a model yields traces that closely reflect real-world execution.

However, this approach has limitations as the resulting execution traces are closely tied to real-world systems. Factors such as the number of NPUs, computation time, and network delays are tightly coupled with the actual system.
Synthetic trace generators can overcome these limitations. In our study, we designed a generative ML model trained with production traces to capture the characteristics of ML traces, such as the relationships between nodes in a trace and their fields. This model can be used to simulate traces of various ML models by varying the number of NPUs. A key advantage of the generative ML model is its ability to obfuscate traces, which is critical for sharing ML traces while protecting intellectual property. Obfuscation is a significant concern when sharing ML traces, and the generative ML model can address this issue by conveying the essential characteristics of the trace while maintaining privacy.

\subsection{Open-source Tools}
Chakra offers a range of open-source tools to help users modify and better understand the execution traces. These tools include the execution trace converter, execution trace visualizer, timeline visualizer, test case generator, and execution trace feeder, all of which are currently available on our GitHub repository\footnote{https://github.com/chakra-et/chakra} and under active development. The execution trace converter is responsible for converting execution traces in various schemas to the Chakra schema. The execution trace visualizer allows users to visualize the dependencies between nodes in a trace. Meanwhile, the timeline visualizer provides a visualization of the execution of nodes when a trace is simulated in a simulator. The test case generator lets users generate arbitrary execution traces by offering libraries to describe traces. The execution trace feeder is a library that other simulators can use to parse traces and feed them into the simulator. Further details will be presented in \autoref{sec:open-source-tools}.

\subsection{Use Cases}
Chakra serves a variety of purposes, including HW-SW co-design, benchmarking, and performance projection. The proposed Chakra schema serves as a unified format for exchanging execution traces, significantly simplifying the process of sharing these traces. Additionally, the generative AI model enables users to produce representative traces based on existing ones. As a result, NPUs or training systems can be optimized using these traces. Furthermore, Chakra can be employed for performance projection, allowing users to estimate the potential performance improvements resulting from investments by modeling systems with varying compute engines or network topologies. A key advantage of Chakra is that it allows companies to use their own internal simulators or toolchains, provided they follow the common format.

%% file: schema.tex
\section{Chakra Schema}
\label{sec:schema}
This section provides a detailed description of the Chakra schema. In this section, we present the goal of the schema and its design. Additionally, we discuss the important design choices made during the development of the Chakra schema.

\subsection{Design Requirements}
\label{sec:chakra-schema-design-requirements}
ML tasks are often presented as graphs, with frameworks such as TensorFlow and PyTorch constructing computational graphs internally to represent model execution~\cite{computational_graphs}. In this study, we introduce \textit{an execution trace} (ET) as a graph that captures the execution details of an ML task. An ET is a trace of ML execution that includes details such as memory accesses, computational load, network communication, and parallelization strategies. While ETs are a useful way to represent an ML task, the structures and metadata of generated ETs may differ depending on the ML framework used. To ensure that the Chakra framework can be used across different ML frameworks, we propose a standard ET schema for performance modeling called Chakra execution trace (\Chakraet).

The Chakra schema was designed to meet the diverse requirements of various teams and organizations. To achieve this goal, we had meetings with many universities and companies to gather input from different teams. We present the requirements that were incorporated into the schema. The first requirement is that the schema should be minimal yet extensible. This is a common requirement across all organizations because they do not want to use up resources on non-used fields, but they also want to extend the schema for their specific purposes. The second requirement is that the schema should be able to model various aspects of ML execution from a performance modeling perspective. While there are many languages and representations designed for presenting ML models, there is no common schema for modeling the performance of ML execution. Therefore, performance modeling is a critical aspect of schema design. The schema should be able to model memory access, computation, and network communication. Additionally, the trace should have dependencies between nodes because many operations have dependencies. The third requirement is that the schema should be able to present execution traces at different stages of execution. If the resulting trace is too tightly coupled to the real system execution, it cannot be used for performance projection or estimation. Therefore, the schema should be able to present execution traces at any level of ML execution.

\begin{table}[t!]
    \centering
    \begin{tabular}{|c|c|c|}
    \hline
    \textbf{Field Name} & \textbf{Data Type}      & \textbf{Description}  \\ \hline
    id                  & required uint64         & Node ID               \\ \hline
    name                & required string         & Node name             \\ \hline
    type                & required enum           & Node type             \\ \hline
    parent              & repeated uint64         & Parent node IDs       \\ \hline
    attribute           & repeated AttributeProto & Any additional fields \\ \hline
    \end{tabular}
    \caption{Chakra node schema.}
    \label{tab:chakra-node-schema}
    
    \begin{tabular}{|c|c|c|}
    \hline
    \textbf{Field Name} & \textbf{Data Type} & \textbf{Description} \\ \hline
    name                & required string    & Name of this attribute    \\ \hline
    type                & required enum      & Type of this attribute   \\ \hline
    doc\_string         & required string    & Description of this attribute    \\ \hline
    f                   & optional float     & Float value        \\ \hline
    i                   & optional int64     & Integer value        \\ \hline
    s                   & optional string    & String value  \\ \hline
    floats              & repeated float     & Repeated float values          \\ \hline
    ints                & repeated int64     & Repeated integer values     \\ \hline
    strings             & repeated string    & Repeated string values  \\ \hline
    \end{tabular}
    \caption{Chakra AttributeProto schema.}
    \label{tab:chakra-attributeproto-schema}
\end{table}

\subsection{Schema}
\label{sec:chakra-schema-detail}
This subsection provides an in-depth look at the Chakra schema, which is presented in \autoref{tab:chakra-node-schema} and \autoref{tab:chakra-attributeproto-schema}. \autoref{tab:chakra-node-schema} outlines the Chakra node schema, while \autoref{tab:chakra-attributeproto-schema} presents the AttributeProto schema. During the design phase, we aimed to create a schema with minimal required fields while making it highly extensible. As demonstrated in \autoref{tab:chakra-node-schema}, the schema only requires a few fields to represent a node, such as id, name, type, parent, and attribute. The id field is used to uniquely identify a node, while the name field is used for naming nodes. 
The nodes in Chakra currently represent compute, memory and communication/networking operations.
The type field, defined as an enum, can be one of the following: INVALID, MEM\_LOAD, MEM\_STORE, COMP, COMM\_SEND, COMM\_RECV, or COMM\_COLL\footnote{SEND and RECV are used for pt-to-pt messages, while COLL for collective messages.}. 
INVALID works as an initial value of the type field, and it is used to ignore non-critical nodes. Memory types are used for tracing and modeling memory systems, and communication types are used for network systems. All additional metadata to model computation, memory access, and network are encoded in the attribute field. The parent field is a repeated field that holds the IDs of parent nodes to encode dependencies between nodes. Finally, the attribute field is a repeated field that significantly improves the schema's extensibility. This idea was inspired by the ONNX schema. \autoref{tab:chakra-attributeproto-schema} shows the schema for AttributeProto, which is essentially a key-value pair. AttributeProto supports six data types that can be identified using the type field: float, int, string, and their repetitions. It should be emphasized that this schema satisfies all of the requirements introduced in \autoref{sec:chakra-schema-design-requirements}.

Please note that this schema satisfies all of the requirements introduced in \autoref{sec:chakra-schema-design-requirements}. First, the schema is designed to be minimal and extensible. The Chakra node schema has critical fields such as id, name, type, parent, and attribute. All other fields can be presented as in the attribute field. However, this design necessitates complex execution trace collection and parsing codes to read and interpret all attribute fields. Second, the schema can model the performance of ML execution. As the type field suggests, nodes in a trace represent memory access, computation, and communication. Their dependencies are presented with the parent field, and various compute engines and networks can be modeled by adding more metadata in the attribute field. Third, the schema can present traces collected at any level because the attribute field is extensible.

\subsection{Discussion}
This subsection delves into the design choices made and their associated implications.

\subsubsection{Pre-execution vs. post-execution}
Determining the appropriate level for trace collection is a crucial design choice. Generally, trace collection levels can be classified as either pre-execution or post-execution. Pre-execution refers to a stage where an ML model has not yet been explicitly optimized or tied to specific system configurations, such as compute engine type, memory bandwidth, or network topology. Collecting a trace at the pre-execution stage allows for performance projection across various systems, as it is not tethered to a particular configuration. Conversely, the post-execution stage involves optimizing and executing the ML model on an actual system. Traces gathered during this stage are tightly linked to the real system on which the model is run. Although these traces accurately represent real-world phenomena, their utility for performance projection is limited due to the varying optimization and parallelization strategies required by different systems. The Chakra schema is designed to accommodate any execution stage. However, how to collect traces from a real system without being tightly bound to it remains an open question.

\subsubsection{Multi-node representation}
In the current Chakra implementation, it is assumed that each NPU possesses a corresponding trace. Given $N$ NPUs, $N$ distinct traces are generated. This design choice offers the advantage of simplifying trace collection and replay, as each NPU does not need to concern itself with the traces or execution of other NPUs. Nonetheless, it constrains the potential for global optimization, wherein a global scheduler could dynamically allocate work to each NPU or redistribute tasks from one NPU to another. Although the Chakra schema does not inherently restrict such an approach, it essentially operates under the assumption that each NPU maintains an individual trace.

\subsubsection{Execution graphs vs. execution traces}
Execution graphs and execution traces are closely related concepts, both represented as graphs, but they serve distinct purposes. Execution graphs are designed to facilitate precise model execution and performance optimization within machine learning frameworks such as PyTorch, TensorFlow, and ONNX. In contrast, execution traces do not focus on the exact execution of models; rather, they concentrate on replaying, simulating, and emulating processes. The primary objective of execution traces is to assess performance and coverage, ideally operating in a framework-agnostic manner.

%% file: generative_model.tex
\section{ET collection and synthesis}
Collecting all production traces data on a regular cadence is feasible with the tooling support provided by Chakra. However, due to the massive amount of data, it is not scalable or effective to replay all of them for benchmarking. Additionally, sharing these traces may not be advised since it would be possible to reconstruct the original model. This could result in proprietary model information being shared widely. Therefore, automatically discovering a representative set of traces is crucial to complete the whole system. To address this issue, we propose a methodology for generating representative traces using generative machine learning models based on production trace data.

Our approach enables efficient generation of realistic-looking production and futuristic traces, which can be used for data-driven decision making. Furthermore, it provides representative traces as the single source of truth that can be shared internally and externally given the stochastic nature of the generative models, which naturally provides an obfuscation of the production data. We start our modeling work from analyzing the communication part of the trace, the same framework can be easily extended to full trace.

The communication part of the traces contains communication collectives across a set of GPUs (ranks). There are a few hard constraints on the traces within the same process group to ensure the replay system can run these traces without raising errors. To better address these constraints, we propose a concept of “master trace”, which is a lossless compression of all $N$ traces of all ranks taking all constraints into consideration. We can construct a master trace from $N$ traces of all ranks, and reconstruct $N$ traces from this master trace without losing any information. The master trace helps us to learn a distribution and synthesize traces without worrying about violating the constraints of the replay system.

\begin{figure}[t!]
    \centering
    \includegraphics[width=0.98\columnwidth]{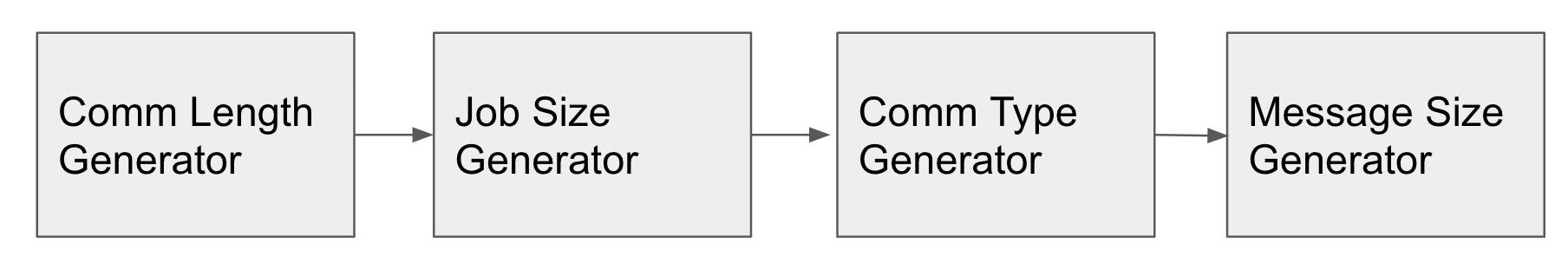}
    \\
    \caption{Generative AI model for ET synthesis.}
    \label{fig:genai_model}
\end{figure}

Based on the master trace data, we developed a hierarchical ML model comprising several generators that ensure the synthesized traces follow the expected data distributions. See \autoref{fig:genai_model} for an overview of this model. There are a couple of advantages for the hierarchical model that facilitates model research and experimentation. It is modularized and flexible. We can substitute the individual generator parts with various algorithms. We can easily extend the model with additional generation stages when needed. This facilitates model research and experimentation. It is also interpretable and easy for debugging. The model is conceptually easy to understand. Even when we use complex algorithms for the generators, we are able to evaluate performance at each stage and identify bottlenecks for improvements. This is a big advantage for model debugging and interpretation compared to a single blackbox model where debugging is difficult. Moreover, the model of generators follow the actual param trace generation logic. It provides us data insights into the param trace generation stages, enabling deeper understanding of param trace patterns.

For network use cases, the composition of collective types and the message sizes are the key factors. The Comm Type Generator ensures that the generated collective types align with the collective composition of production traces. We have identified two major clusters of collective type compositions. The Message Size Generator ensures that the message size generated follows the distribution of each collective type. We developed Gaussian mixture models to fit the data and draw samples from the fitted distribution. The synthesized samples align relatively well with the true distribution. We have modeled the distributions of comm types, comm sequence length, GPU sizes, message sizes, autocorrelation between comm types and message sizes, and message splits across GPUs, as well as the dependencies between the generators. We have tested the synthesized traces using the replay system and found that they work correctly on our training cluster. 

The generative models can be extended to generate traces for larger futuristic system scales and compute \& communication traces, which will capture compute-communication overlap, dependency structure, etc. Overall, our generative AI models enables efficient generation of representative traces, which will be a valuable tool for optimizing distributed AI training and evaluating future hardware designs.

%% file: open_source_tools.tex
\section{Open-source Tools}
\label{sec:open-source-tools}
In this section, we introduce the open-source tools offered by the Chakra project, which are currently under active development. More tools can be added in the future.

\subsection{Execution Trace Converter}
\label{sec:execution-trace-converter}
While various ML frameworks allow users to collect execution traces of ML models, they are often in different formats, hindering the exchange of traces between different organizations. The ET converter's role is to translate ETs from various schemas into the Chakra schema. Currently, the converter supports PyTorch and FlexFlow~\cite{jia2019beyond}, and considering that FlexFlow supports Keras, PyTorch, and ONNX, it can support most ML frameworks. TensorFlow support is planned for future work.

The ET converter has prior knowledge of the input execution trace schema and can extract Chakra fields from the input file. For instance, PyTorch execution traces are in the JSON format, and the converter uses the JSON parser to read the execution traces. In PyTorch, computation nodes have the duration field if computation time is encoded after collecting traces, allowing computation nodes to be identified. Communication nodes in PyTorch are identified by the name of nodes, which is "record\_param\_comms". The communication type and size are determined by its child node, whose name starts with "nccl:". Unlike PyTorch traces, FlexFlow traces are in the graphviz format, and the FlexFlow codes needed to be slightly modified to identify node types and convey more metadata. Node types in FlexFlow are identified with the name field, and each node has corresponding metadata fields. One significant difference between FlexFlow and PyTorch is that collective communications in FlexFlow are broken down into multiple peer-to-peer data transfers. If the converter encounters uninterpretable nodes, they are classified as invalid nodes and ignored in other toolchains. It is important to note that in Chakra, each NPU generates a separate trace. Therefore, if an input file describes the behavior of all NPUs, the converter must split it into multiple pieces correctly.

\subsection{Execution Trace Visualizer}
\label{sec:execution-trace-visualizer}
\begin{figure}[t!]
    \centering
    \includegraphics[width=0.98\columnwidth]{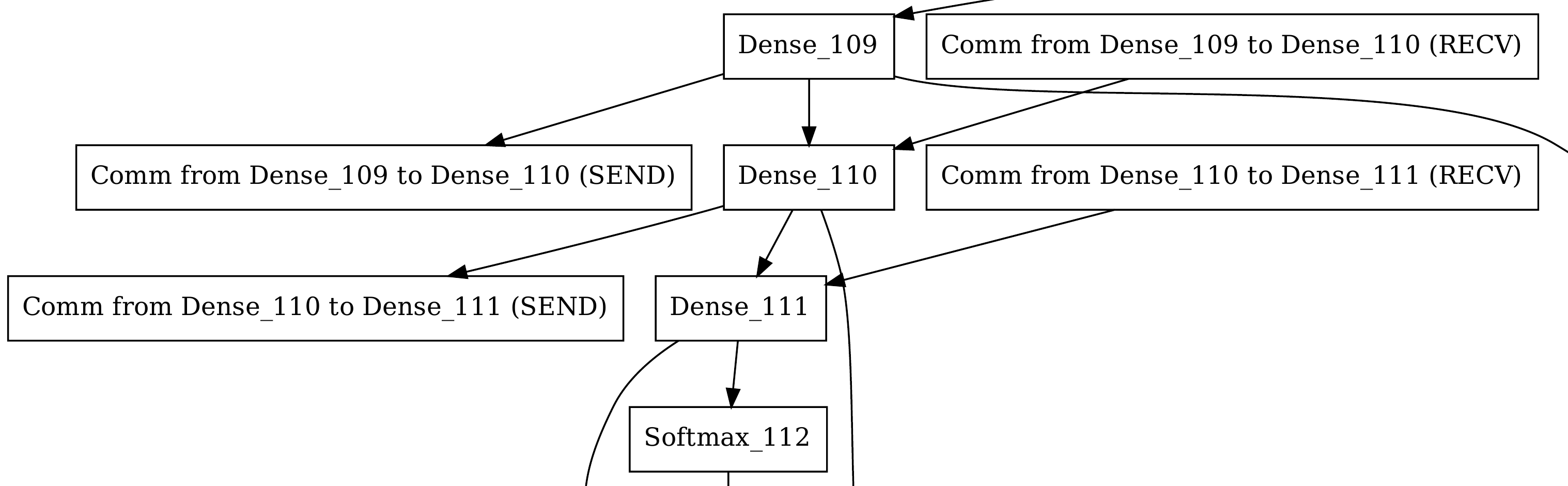}
    \\
    \caption{Execution trace visualization example.}
    \label{fig:execution_trace_visualization_example}
\end{figure}

The execution trace visualizer is designed to visualize execution traces in the Chakra schema. The visualizer's input file should conform to the Chakra schema, and it can output files in either the graphviz or PDF format. The visualizer is a helpful tool for researchers to understand the structure of execution traces. Moreover, it can be used for debugging simulators or emulators that take Chakra execution traces. By default, the visualizer encodes the names of tasks and dependencies between tasks. Users can easily modify the visualizer to encode additional metadata such as compute time and communication size. An example of a visualization from the execution trace visualizer is presented in \autoref{fig:execution_trace_visualization_example}.

\subsection{Execution Timeline Visualizer}
\label{sec:execution-timeline-visualizer}
\begin{figure}[t!]
    \centering
    \includegraphics[width=\columnwidth]{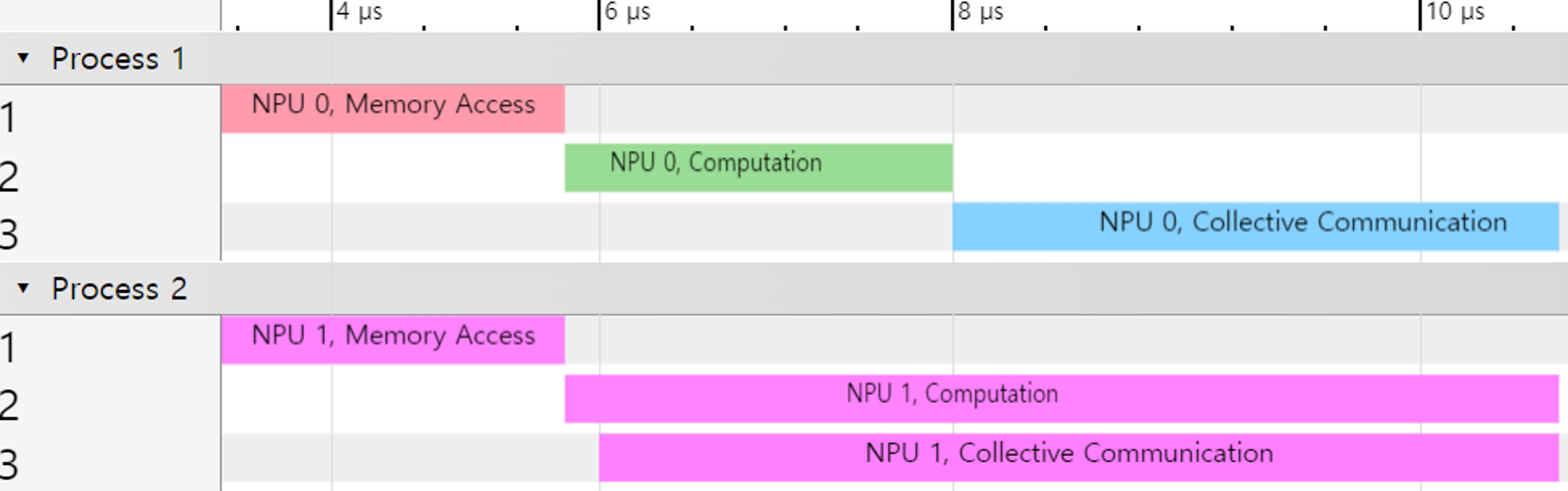}
    \\
    \caption{Execution timeline visualization example.}
    \label{fig:execution_timeline_visualization_example}
\end{figure}

The execution timeline visualizer presents task execution on each NPU in a timeline, taking a CSV file describing the task execution in a callback-based simulator. An example input format for the visualizer is presented in Snippet~\ref{snippet:timeline-visualizer-input-format}. The output timeline can be visualized with the trace event profiling tool in Chrome (chrome://tracing). \autoref{fig:execution_timeline_visualization_example} shows a screenshot of the visualization result, where tasks running on each NPU are presented as bars on the timeline, with their start and finish times. Tasks on an NPU are grouped as a process, and are classified into several sub-groups depending on the task type. In this example, two NPUs are shown, and memory access, computation, and communication have TID 1, 2, and 3, respectively. This tool allows researchers to easily identify bottlenecks in distributed ML tasks. For instance, in NPU 1, communication time is not overlapped with computation, unlike in NPU 2. Therefore, the task in NPU 1 can be optimized by overlapping communication with computation.

\begin{figure}[!t]
\begin{minipage}{\linewidth}
\begin{lstlisting}[basicstyle=\footnotesize]
issue,[gpu_id],[curr_cycle],[node_id],[node_name]
callback,[gpu_id],[curr_cycle],[node_id],[node_name]
issue,[gpu_id],[curr_cycle],[node_id],[node_name]
issue,[gpu_id],[curr_cycle],[node_id],[node_name]
callback,[gpu_id],[curr_cycle],[node_id],[node_name]
callback,[gpu_id],[curr_cycle],[node_id],[node_name]
...
\end{lstlisting}
\vspace{-1em}
\captionof{lstlisting}{Execution timeline visualizer input format.}
\label{snippet:timeline-visualizer-input-format}
\end{minipage}
\end{figure}

\subsection{Test Case Generator}
While execution traces are expressive and can effectively encode real-world ML traces, their limited readability poses challenges for implementing ET-based tools. The test case generator allows users to define arbitrary execution traces that can subsequently be used to generate test cases. Additionally, the test case generator allows manual modeling of ML models. The test trace generator offers an array of C++ classes for composing Chakra execution traces. Snippet~\ref{snippet:synthetic-trace-generator-example} demonstrates how an execution trace can be defined within the tool. As illustrated in the code snippet, users can create a new trace and designate the dependencies between nodes. We also offer several default trace examples, encompassing data parallel, model parallel, and pipeline parallel instances for a synthetic ML model.

\subsection{Trace Feeder for Simulators}
\label{sec:execution-trace-feeder}
We offer a trace feeder library that is designed for simulators. One of the use cases for Chakra is simulating distributed ML systems for performance projection or what-if analysis. We anticipate that numerous companies and research institutions will actively share traces in the Chakra schema and use them with their simulators, whether open-source such as ASTRA-sim~\cite{rashidi2020astra, astrasim2} or SST~\cite{rodrigues2011structural}, or proprietary ones. The trace feeder library is implemented in C++, eliminating the need for users to repeatedly implement the parsing code. The trace feeder library significantly reduces users' engineering overhead. The public member functions of the trace feeder class are presented in Snippet~\ref{snippet:trace-feeder-class-definition}. With these functions, users can create a trace feeder instance and retrieve nodes or resolve dependencies.

\begin{figure}[!t]
\begin{minipage}{\linewidth}
\begin{lstlisting}[basicstyle=\footnotesize]
et = new DependencyGraph(getFilename(..., npu_id));

node1 = et->addNode(ChakraProtoMsg::COMP_NODE);
node1->set_name("COMP_NODE");
node1->set_simulated_run_time(5);

node2 = et->addNode(ChakraProtoMsg::COMP_NODE);
node2->set_name("COMP_NODE");
node2->set_simulated_run_time(5);

et->assignDep(node1, node2);
\end{lstlisting}
\captionof{lstlisting}{Test case generator codes.}
\label{snippet:synthetic-trace-generator-example}
\end{minipage}
\end{figure}

\begin{figure}[!t]
\begin{minipage}{\linewidth}
\begin{lstlisting}[basicstyle=\footnotesize]
void addNode(std::shared_ptr<EGFeederNode> node);
void removeNode(uint64_t node_id);
bool hasNodesToIssue();
std::shared_ptr<EGFeederNode> getNextIssuableNode();
void pushBackIssuableNode(uint64_t node_id);
std::shared_ptr<EGFeederNode> lookupNode(uint64_t node_id);
void freeChildrenNodes(uint64_t node_id);
\end{lstlisting}
\captionof{lstlisting}{Trace feeder class public member functions.}
\label{snippet:trace-feeder-class-definition}
\end{minipage}
\end{figure}

%% file: use_cases.tex
\section{Use Cases}
\label{sec:use-cases}

\subsection{Replay Benchmarks}
Traditional benchmark creation involves starting with choosing representative applications and adapting and/or simplifying their implementation to highlight a few specific behaviors. On the contrary, detailed operator and dependency information from Chakra ETs can be used to develop \emph{replay benchmarks}. These benchmarks essentially replay the graph to mimic the application behavior. Chakra ET replay benchmarks offer the following unique advantages. First, ET replay drastically reduces the benchmark software dependency (e.g. various libraries a full application usually comes with, many of them do not have a direct relationship to the performance evaluation, but rather a vestige of the original application's other business logic). This makes the resulting benchmark agile and easy to deploy. Second, creating/adapting an application to benchmarks requires a lot of efforts; hence, creating benchmarks directly from ET collected during the normal fleet operation can save a tremendous amount of work. Lastly, because it is possible to build an automatic benchmark creation pipeline continuously, either sourcing from the fleet workloads or through the fore-mentioned generative AI techniques, we can keep our benchmark suites up-to-date at scale. 

PARAM Comms Replay benchmarks replayed collective operations from a trace. More recently, PARAM Full Replay benchmarks (Mystique) ~\cite{liang2023mystique} successfully used PyTorch ETs (with both compute and collective operations) to generate benchmarks for AI workloads in production. It proposed an automated end-to-end infrastructure to collect and generate benchmarks from the production fleet and demonstrated high accuracy in terms of both performance and system-level metrics. By standardizing the ET schema with Chakra, we can extend this methodology to a wide range of sources; not just ETs collected from PyTorch. We are currently extending the PARAM replay benchmarks to use Chakra ET schema.

\subsection{Simulation for Performance Projection}
We demonstrate the value of Chakra in simulating the performance of ML training tasks on future systems through two case studies: scaling the number of NPUs and network bandwidth. Our proof-of-concept converts PyTorch ETs to Chakra ETs and uses this to drive an open-source training system simulator (ASTRA-sim). 

\subsubsection{Implementation.}
\begin{figure}[t!]
    \centering
    \includegraphics[width=\columnwidth]{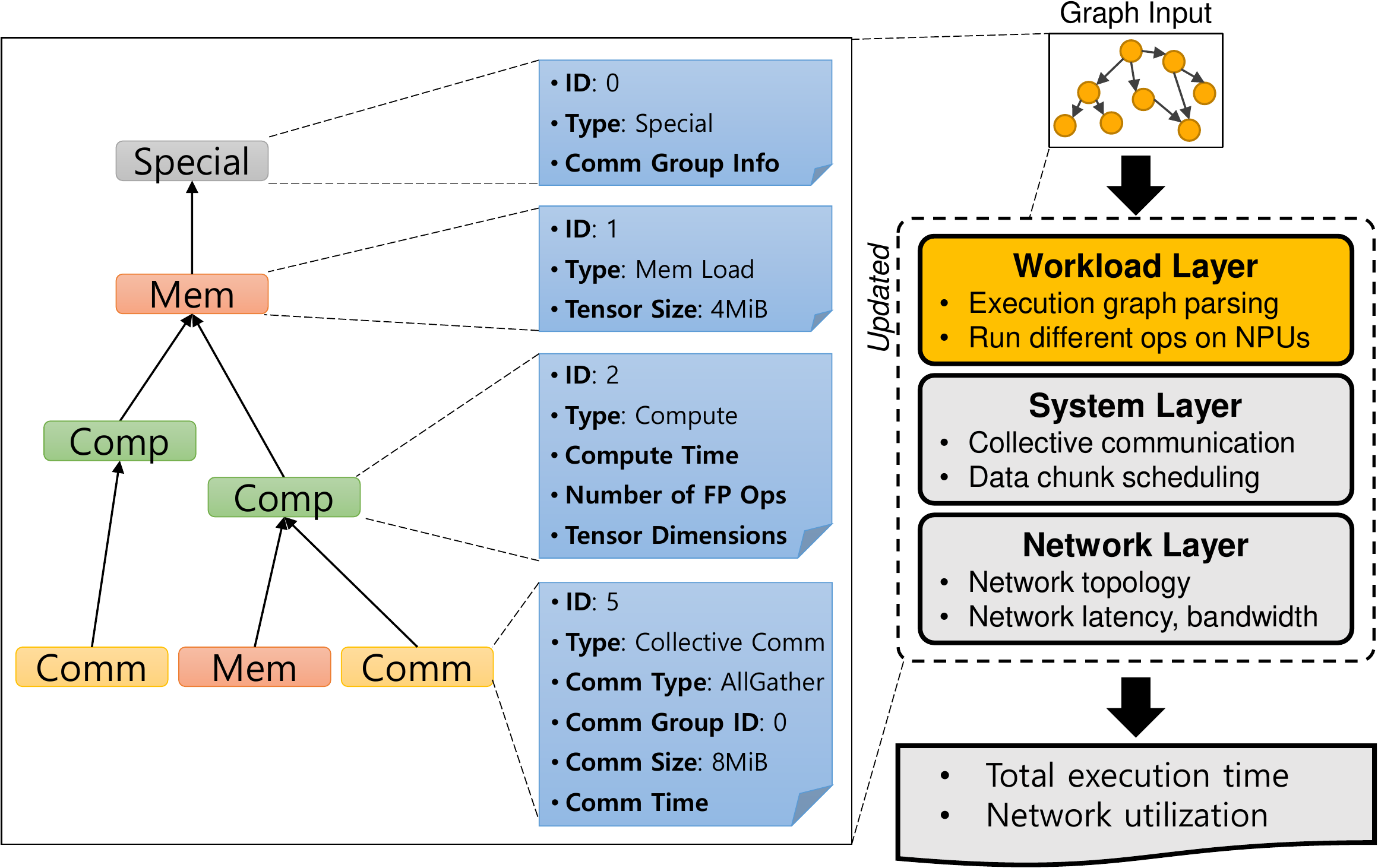}
    \\
    \caption{Overview of ASTRA-sim-v2. An execution trace is shown on the left and ASTRA-sim-v2 is shown on the right. Unchanged software layers from ASTRA-sim-v1 are presented in gray, and the updated layer is highlighted with yellow.}
    \label{fig:astra_sim_overview}
\end{figure}

The Chakra infrastructure enables users to utilize any Chakra-compatible simulators for performance modeling. In this paper, we modify ASTRA-sim~\cite{rashidi2020astra}, a distributed ML system simulator, to make it Chakra-compatible. We choose ASTRA-sim because it models the HW-SW co-design space of training systems, such as workload parallelization, collective communication and scheduling, and networking (\autoref{fig:astra_sim_overview}), and has been validated against real systems.

We denote the original ASTRA-sim as ASTRA-sim-v1 and the updated ASTRA-sim as ASTRA-sim-v2. \autoref{fig:astra_sim_overview} presents the key changes made to ASTRA-sim-v2 to support ETs. First, we update the workload layer to read \Chakraet as its input file and execute the graph, replacing v1's text-based input and manually implemented training loops. Second, we add support for NPUs to execute different operations simultaneously~\cite{communicator_group}, unlike v1 that requires the same operations on each NPU. The graph-input workload layer issues dependency-free nodes from the ET one by one, and an issued node completes after a specific number of cycles. The number of cycles to simulate a compute or communication node can either be read directly from the ET or simulated via ASTRA-sim-v2, which models both NPU's computation capability and network latency/bandwidth. Compute and communication time read directly from the ET effectively models the real system on which the ET was collected, while simulation can be used for modeling future systems\footnote{For instance, a SW company can study the impact of changing the network topology with current NPUs by using the NPU runtime from the ET and simulating the network using an internal Chakra-compatible simulator; A HW vendor can study the impact of their next-generation NPUs while reading communication times from the ET.}. Once a node is completed, dependent nodes become dependency-free if they do not have any other dependencies. The graph-based execution engine repeats this process until all nodes are consumed.

\subsubsection{Target Systems}
We evaluate two network topologies in this experiment: 2D-torus and DGX2. The 2D-torus is an 8$\times$8 2D-torus topology with 64 NPUs and a link bandwidth of 62GB/s. In contrast, the DGX2 is a hierarchical network topology connected with switches, with network bandwidths for the first and second dimensions set to 600GB/s and 37.5GB/s, respectively.

\subsubsection{Target ML Training Tasks}
We utilize three models for this study: \texttt{MLP}, \texttt{DLRM}, and \texttt{Transformer}. The \texttt{MLP} model is a synthetic workload that consists of six layers. For the \texttt{MLP}, we implement model-parallel (MP) and two hybrid parallel schemes: data parallel (DP) along the first network dimension and MP along the second (referred to as \texttt{MLP-DP-MP}), and the reverse (\texttt{MLP-MP-DP}). The \texttt{Transformer} model has 165M parameters and employs hybrid MP-DP using ZeRO-2 optimization for the DP dimension. Meanwhile, the \texttt{DLRM} model applies MP for its embedding layers and DP for the MLP layers. The ETs for \texttt{MLP} are generated synthetically, while the ETs for \texttt{Transformer} and \texttt{DLRM} are modeled based on real-world models and converted into the Chakra schema by the converter (\autoref{sec:execution-trace-converter}).

\begin{figure}[t!]
    \centering
    \begin{subfigure}[b]{\columnwidth}
        \centering
        \includegraphics[width=\textwidth]{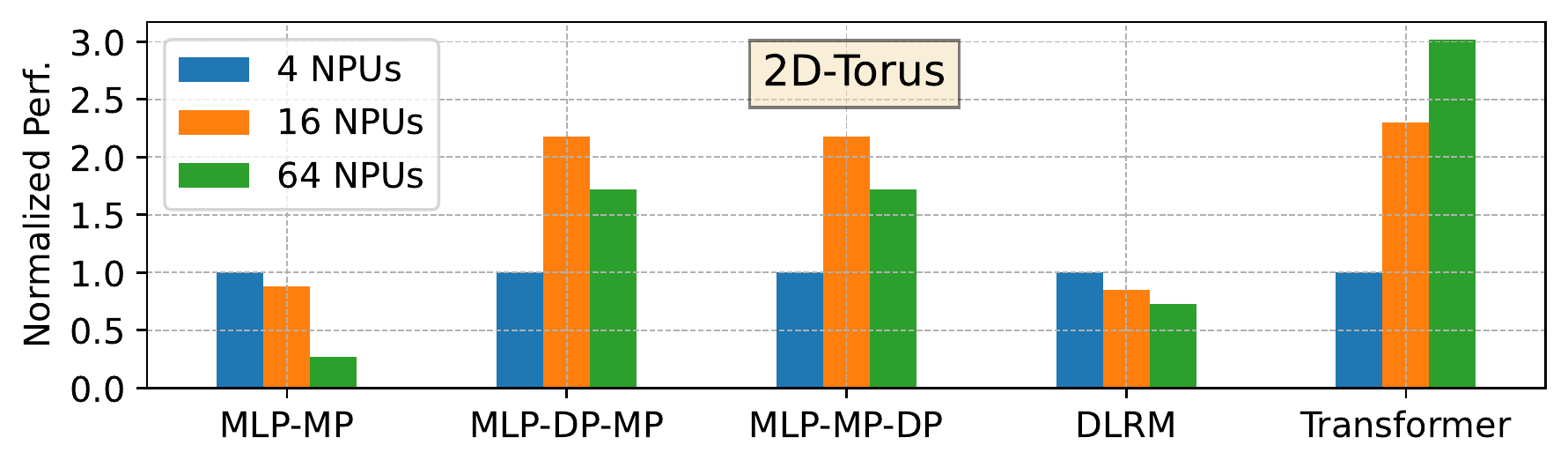}
    \end{subfigure}
    \\
    \begin{subfigure}[b]{\columnwidth}
        \centering
        \includegraphics[width=\columnwidth]{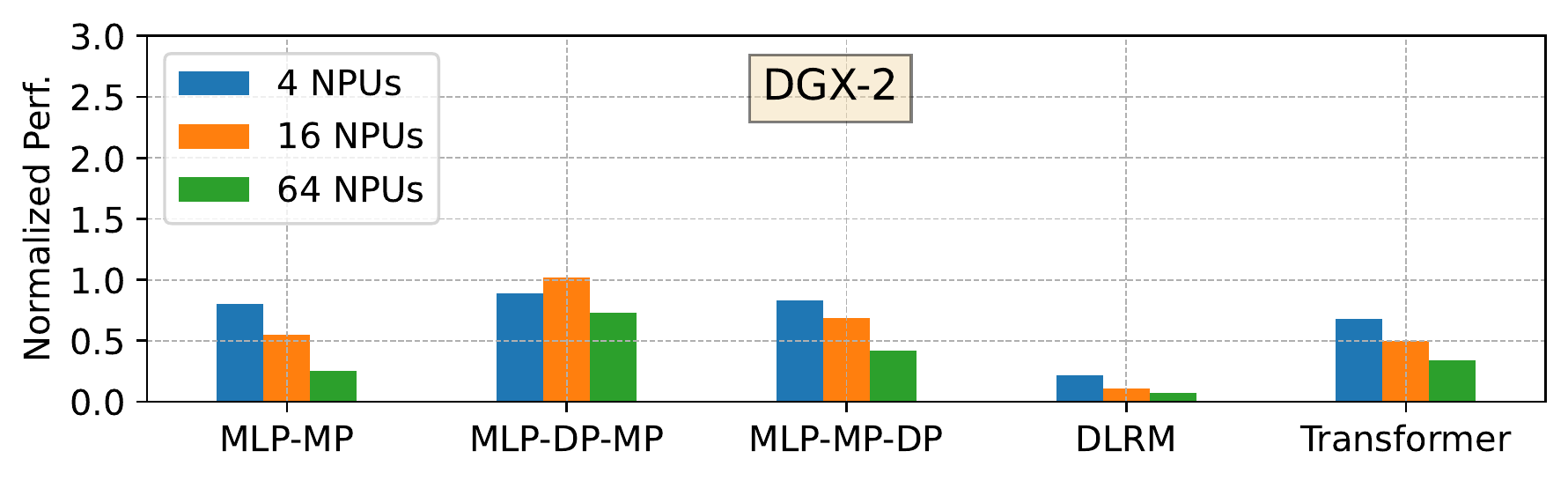}
    \end{subfigure}
    \\
    \caption{Performance of training tasks while varying the number of NPUs.}
    \label{fig:what-if-number-of-npus}
\end{figure}

\begin{figure}[t!]
    \centering
    \begin{subfigure}[b]{\columnwidth}
        \centering
        \includegraphics[width=\columnwidth]{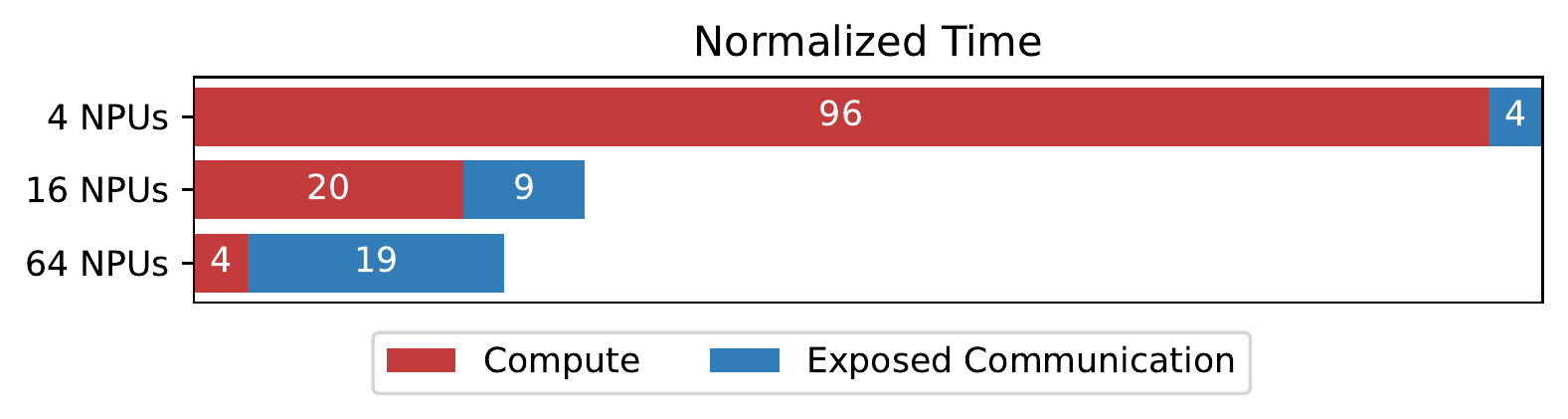}
        \\
        \caption{\texttt{Transformer} on 2D-torus}
    \end{subfigure}
    \\
    \begin{subfigure}[b]{\columnwidth}
        \centering
        \includegraphics[width=\columnwidth]{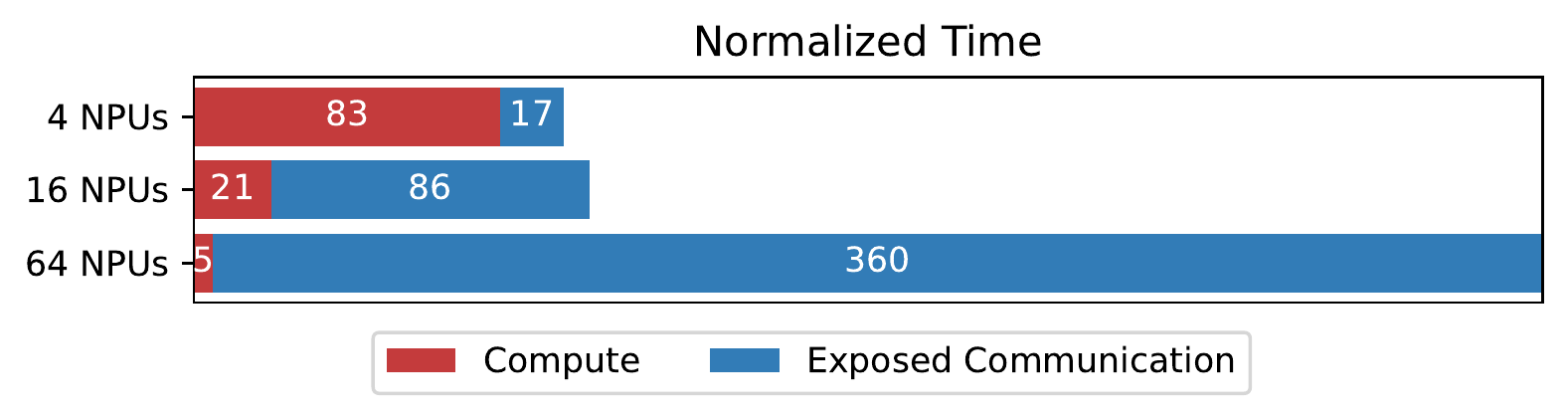}
        \\
        \caption{\texttt{MLP-MP} on 2D-torus}
    \end{subfigure}
    \\
    \caption{Execution time breakdown of training tasks on 2D-torus while varying the number of NPUs.}
    \label{fig:execution-time-breakdown}
\end{figure}

\subsubsection{Experiment Results}
\circled{1}
\textbf{Scaling the Number of NPUs.}
\autoref{fig:what-if-number-of-npus} displays the normalized runtime of workloads while varying the number of NPUs between 4, 16, and 64. Performance is defined as the inverse of execution cycles, with values normalized to the 2D-torus topology with four NPUs. Interestingly, performance may not always improve with an increase in the number of NPUs. The only case where performance consistently improves is with \texttt{Transformer} on the 2D-torus topology. As a compute-dominant model, \texttt{Transformer}'s total compute time is 25 times the total communication time on the 2D-torus topology with four NPUs. Therefore, increasing the number of NPUs can considerably enhance \texttt{Transformer}'s performance by effectively reducing the compute time. \autoref{fig:execution-time-breakdown} presents the breakdown of normalized execution time for each configuration, with a bar representing the sum of total compute time and exposed communication time, normalized to the 4-NPU configuration. Exposed communication time is defined as the communication time that does not overlap with computation. In \texttt{Transformer}, while increasing the number of NPUs leads to increased exposed communication time, the reduction in compute time is significant. However, in \texttt{MLP-MP}, exposed communication time dominates and eventually hinders performance when the number of NPUs is increased.

\begin{figure}[t!]
    \centering
    \begin{subfigure}[b]{\columnwidth}
        \centering
        \includegraphics[width=\textwidth]{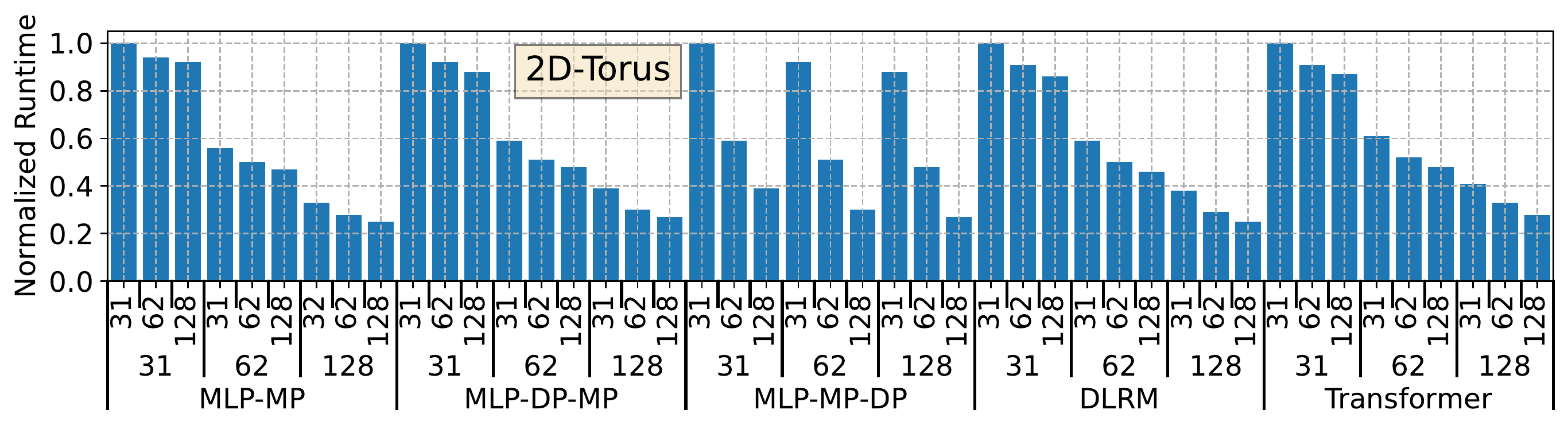}
    \end{subfigure}
    \\
    \begin{subfigure}[b]{\columnwidth}
        \centering
        \includegraphics[width=\columnwidth]{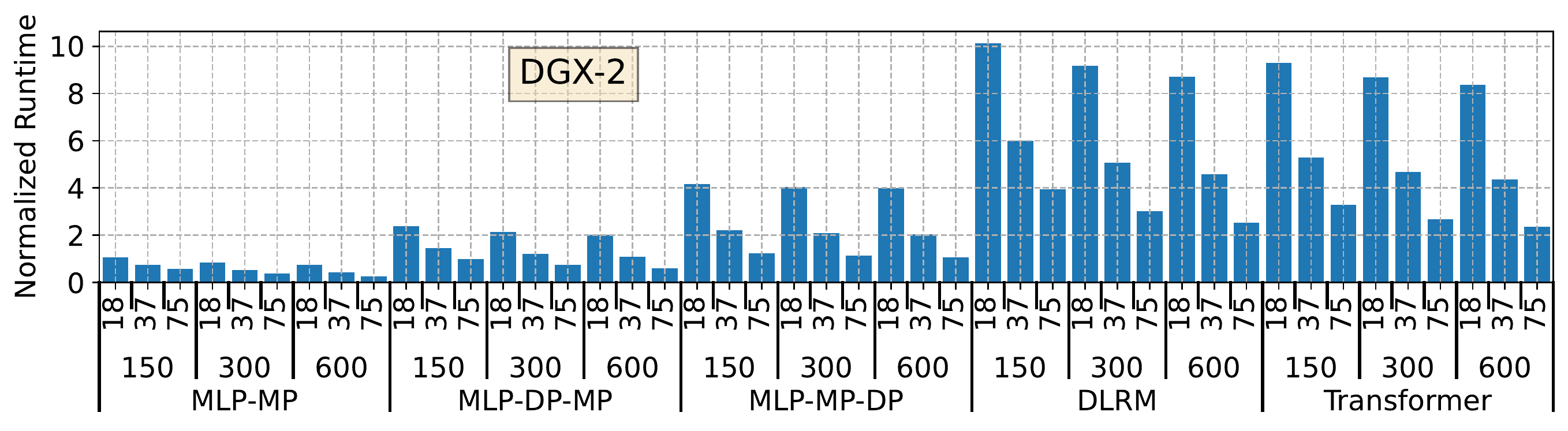}
    \end{subfigure}
    \\
    \caption{Runtime of training tasks with varying network bandwidth (GBps).}
    \label{fig:what-if-network-bandwidth}
\end{figure}

\circled{2} \textbf{Varying Network Bandwidth.}
In this experiment, we measure the normalized runtime of training loops while varying network bandwidths, with runtime normalized to the 2D-torus network and the lowest network bandwidth combination. The number of NPUs is set to 64. In the 2D-torus topology, we vary the network bandwidth of the first dimension (\texttt{dim1}) and the second dimension (\texttt{dim2}) between 31, 62, and 128GB/s. In the DGX2-like topology, we vary the dim1 network bandwidth between 150, 300, and 600GB/s, and the dim2 network bandwidth between 18.75, 37.5, and 75GB/s, respectively. \autoref{fig:what-if-network-bandwidth} illustrates the normalized runtime of workloads while varying the network bandwidth of the first and second dimensions, with dim2 bandwidths displayed on the first row of the xticklabel and dim1 bandwidths on the second row. Generally, training tasks exhibit better performance on the 2D-torus topology compared to the DGX2-like topology. Among the models, \texttt{MLP-MP} is the most sensitive training task, with its high sensitivity to network bandwidths stemming from the dominant exposed communication time. On the 2D-torus topology with the lowest bandwidth combination, the exposed communication time of \texttt{MLP-MP} is 128 times the total computation time.

\subsection{Additional Use Cases}
Beside simulations, we can model a ML workload's component or end-to-end performance behavior by combining execution traces with profiling data from real hardware. This is especially useful in large scale ML infrastructures where thousands of models with different complexity and behaviors are being trained daily. GPUs have been widely used as accelerators for large scale ML model training. A number of additional use cases have successfully applied this technology to model GPU performance and solve these problems:
\begin{enumerate}
    \item Estimate operator and communication cost for embedding table sharding and load-balancing in distributed training~\cite{zha2022dreamshard, zha2022autoshard}.
    \item Iteration latency estimation for performance tuning and speed-of-light modeling~\cite{lin2022building}.
\end{enumerate}
Cost modeling is an important step in the embedding table sharding algorithm. The Dreamshard~\cite{zha2022dreamshard} framework utilizes data collected from traces to guide the computation and communication micro-benchmarks to collect component level performance data. The benchmark results are used to train neural cost models. Millions of different operator configurations are measured through PARAM benchmarks. The trained cost model enabled Dreamshard to outperform the baselines on all the tasks. 

To extend beyond component level modeling, the GPU performance model~\cite{lin2022building} used a critical-path-based algorithm to predict the per-batch training time of ML models by traversing its execution trace. It achieved less than 10\% geometric mean average error (GMAE) in all kernel performance modeling, and 4.61\% and 7.96\% geomean errors for GPU active time and overall E2E training latency prediction including overheads.

The execution trace also provides crucial information for optimization analysis. Operator out-of-order execution via data dependency analysis, combined with data I/O movement and communication modeling, it's now feasible to measure and identify optimization opportunities across thousands of ML workloads at scale.

%% file: related_work.tex
\section{Related Work}
\label{sec:related_work}
The technique of collecting traces to understand system internals and identify bottlenecks is widely adopted and deployed. Google, for instance, collected and released instruction traces from its servers for representative cloud workloads~\cite{google_workload_traces}. Chakra is a similar effort aimed at ML systems research, specifically for distributed ML systems. In Chakra, we proposed a common schema for ML execution traces.

As ML tasks are often represented as graphs, numerous graph schemas have been proposed. ONNX is one of the most popular graph schemas, designed to facilitate the exchange of models between different ML frameworks. Although ONNX shares similarities with Chakra, such as enabling data exchange between different teams, the objectives of the two differ. ONNX focuses on the exchange of \textit{models} between various frameworks, while Chakra's primary goal is to exchange \textit{execution traces} between different teams. 

When collected in the pre-execution stage, traces closely resemble the model itself, albeit with additional annotations. The traces collected at the pre-execution stage are similar to the graphs presented in prior autotuning studies, which aim to optimize the parallelism strategies of ML models~\cite{wang2019parallel, santhanam2021distir, schaarschmidt2021automap}. Unity is a representative study in this approach. It optimizes ML models by simultaneously applying algebraic transformations and parallelization. To perform optimizations, Unity defines a new graph schema called the parallel computation graph.

%% file: conclusion.tex
\section{Conclusion}
\label{sec:conclusion}
In this paper, we propose Chakra, an open graph schema for standardizing workload specification capturing key operations and dependencies, aka Execution Trace (ET). In addition, we propose a complementary set of tools/capabilities to enable collection, generation, and adoption of Chakra ETs by a wide range of simulators, emulators, and benchmarks. We believe such a framework would enable wider co-optimization of current AI systems and co-design of future systems for AI inference and training. We solicit feedback from readers and are interested in developing productive collaborations to help build an open industry-wide ecosystem around Chakra schema/tools.

%% file: acknowledgements.tex
\section{Acknowledgements}
The authors would like to thank Winston Liu, Dan Mihailescu, and Andy Balogh from Ixia/Keysight Technologies for their early feedback and for engaging in insightful discussions that guided the development of the Chakra ET schema and tools. The authors would like to thank Mingyu Liang and Zhongyi Lin for discussions on PARAM replay benchmarking use case. This work was supported, in part, by Semiconductor Research Corporation.